\documentclass[10pt,twocolumn,letterpaper]{article}

\usepackage{cvpr}
\usepackage{times}
\usepackage{epsfig}
\usepackage{graphicx}
\usepackage{amsmath}
\usepackage{amssymb}

\usepackage[breaklinks=true,bookmarks=false]{hyperref}

\cvprfinalcopy 

\def\cvprPaperID{****} 

\begin{document}

\title{Out of Distribution Detection on ImageNet-O}

\author{Anugya Srivastava\\
Courant, NYU\\
{\tt\small as14770@nyu.edu}
\and
Shriya Jain\\
Courant, NYU\\
{\tt\small sj3409@nyu.edu}
\and
Mugdha Thigle\\
Courant, NYU\\
{\tt\small mht5820@nyu.edu}
}
\maketitle

\begin{abstract}
    Out of distribution (OOD) detection is a crucial part of making machine learning systems robust. The ImageNet-O dataset \cite{Hendrycks_2021_CVPR} is an important tool in testing the robustness of ImageNet \cite{5206848} trained deep neural networks that are widely used across a variety of systems and applications. Inspired by \cite{tajwar2021true}, we aim to perform a comparative analysis of OOD detection methods on ImageNet-O \cite{Hendrycks_2021_CVPR}, a first of its kind dataset with a label distribution different than that of ImageNet, that has been created to aid research in OOD detection for ImageNet \cite{5206848} models. As this dataset is fairly new, we aim to provide a comprehensive benchmarking of some of the current state of the art OOD detection methods on this novel dataset. This benchmarking covers a variety of model architectures, settings where we haves prior access to the OOD data versus when we don't, predictive score based approaches, deep generative approaches to OOD detection, and more. The code is available \href{https://github.com/anugyas/OOD-on-NAE}{here.}
\end{abstract}
\section{Introduction}
Generally, machine learning (ML) and deep learning (DL) based approaches rely on the assumption that the train set and test set come from the same data distribution. However, when these approaches are deployed in the real world, this assumption becomes a very strong one and may not hold, more often than not. This makes the inferences made by the ML/DL model for this out of distribution data unreliable and can have far reaching consequences. Given this level of impact in various aspects of technology and life, detecting out of distribution data is imperative in making these ML/DL systems robust and reliable to use. Moreover, the prevalence of ImageNet \cite{5206848} trained models in these systems makes it all the more important to identify scenarios where we're using ImageNet \cite{5206848} models on data that comes from a data distribution than that of ImageNet \cite{5206848}. This is where ImageNet-O \cite{Hendrycks_2021_CVPR} can be instrumental in helping develop techniques in identifying such scenarios, and is one of the first of its kind that has been developed to aid research for OOD detection for ImageNet \cite{5206848} models.\\
Our contribution has been in benchmarking various widely used OOD detection techniques on this novel dataset i.e. ImageNet-O \cite{Hendrycks_2021_CVPR} and providing analysis for the same. As there are a plethora of OOD approaches out there [\cite{DBLP:journals/corr/HendrycksG16c}, \cite{DBLP:journals/corr/abs-2010-03759}, \cite{DBLP:journals/corr/abs-1812-04606},\cite{DBLP:journals/corr/LiangLS17}, \cite{lee2018simple}, \cite{thulasidasan2021effective}, \cite{ren2021simple}, \cite{kirichenko2020normalizing}, \cite{hendrycks2019deep}] and many more.\\
We have included the predictive score based approach of \textbf{Maximum Softmax Probability (MSP)} \cite{DBLP:journals/corr/HendrycksG16c} that is widely used across OOD detection literature as the baseline. The other approaches that we have included are ones that have proposed a novel improvement to existing OOD detection techniques and have even been established as state of the art. These include \textbf{ODIN} \cite{DBLP:journals/corr/LiangLS17}, \textbf{Energy-based OOD Detection} \cite{liu2021energybased} and \textbf{Normalizing Flows} \cite{kirichenko2020normalizing}. Thus, we have 2 predictive score based approaches - MSP \cite{DBLP:journals/corr/HendrycksG16c} and ODIN \cite{liang2020enhancing}, an energy score based approach \cite{liu2021energybased} as well as a deep generative approach \cite{kirichenko2020normalizing} to OOD detection. We detail the methods as well as the various experiment settings we tried in the following sections.\\
As the OOD detection setting that we are looking at is one where ImageNet \cite{5206848} is the in-distribution dataset and ImageNet-O is the out-distribution dataset, we have not included distance based approaches like Mahalanobis distance[\cite{lee2018simple},\cite{ren2021simple}] and Pairwise OOD detection OOD \cite{tajwar2021true}, as they required working with the entirety of the ImageNet \cite{5206848} train set, and proved to be a computational bottleneck.\\
In summary, we present a detailed benchmarking of various OOD detection techniques on the newly released ImageNet-O \cite{Hendrycks_2021_CVPR}, with ImageNet \cite{5206848} being the in-distribution dataset, and provide our analysis for the same. 
\section{Related Work}
Most of the OOD work so far has been limited to using CIFAR as the in-distribution dataset. \cite{hendrycks2018baseline} uses CIFAR-10, CIFAR-100 and MNIST as in distribution and for out of distribution (negative) examples, they use realistic images from the Scene UNderstanding dataset (SUN) and noise. Apart from CIFAR-10 and CIFAR-100, \cite{ren2021simple} uses  Genomics OOD benchmark and CLINC Intent OOD benchmark with a pre-trained BERT model. ODIN \cite{DBLP:journals/corr/LiangLS17} also uses CIFAR-10 as in distribution and test against TinyImageNet as OOD with a  pre-trained DenseNet. \cite{kirichenko2020normalizing} worked with ImageNet as in-distribution and CelebA as OOD. We would like to add to this large body of existing work with a new in-distribution, out-distribution setting which we think is more representative of real world data scenarios. \\
There have been other works on OOD Detection using different settings. Methods like density estimation, clustering analysis and nearest neighbor have been used for detecting low-dimensional out of-distribution. These techniques are unreliable for high dimensional spacecontact@dair-institute.org like image space. In recent years, deep generative models are being used to make OOD detectors like in \cite{schlegl2017unsupervised} where they trained a generative adversarial networks to detect out-of-distribution examples in clinical scenario and \cite{sabokrou2017deepanomaly} where they trained convolutional network to detect anomalies in scenes. There are fair share of challenges while using Deep generative models. Training and optimizing the models is highly difficult especially with large and complex datasets. DGM also assign high likelihood to out-of-distribution data \cite{nalisnick2019deep}. Better models can be created by using modified and improved metrics including likelihood ratio\cite{ren2019likelihood}. There have been many alternative methods that have come up that uses DGM in a modified manner like \cite{serra2020input} and \cite{schirrmeister2020understanding}.
\section{Methods}
Prior works like \cite{Bishop94noveltydetection} and \cite{zhang2021understanding}  state that OOD detection can be formalized as the task of identifying points that lie with a low likelihood under the training distribution, estimated via a model. Thus OOD detection can be viewed as a binary classification problem, with the two classes being out-distribution or not (or vice-versa with in-distribution as the positive label). Please note that in \cite{Hendrycks_2021_CVPR} the out-distribution class is positive whereas the in-distribution class is negative, and we follow the same notation. Moreover, as shown in \cite{tajwar2021true} and \cite{zhang2021understanding}, it is important to be explicit about what in-distribution and out-distribution we are considering while performing and evaluating OOD detection. For our experiments,\\
\textbf{In Distribution Dataset: } ImageNet-1K \cite{5206848} or in some cases TinyImageNet\cite{TinyImageNet} for computational gains. We specify for each method and experiment, the specific in-distribution in consideration.\\
\textbf{Out Distribution Dataset: } ImageNet-O \cite{Hendrycks_2021_CVPR}. This dataset contains 2000 images from 200 classes that contain semantic information differing from the ImageNet-1K \cite{5206848} classes. As seen in Figure 1, there is a label distribution shift between ImageNet-1K \cite{5206848} classes and the ImageNet-O \cite{Hendrycks_2021_CVPR} classes, albeit being visually similar.This is a more real world OOD scenario and hence makes it an important dataset to benchmark on.
\begin{figure}[t]
\includegraphics[width=8cm, height=6cm]{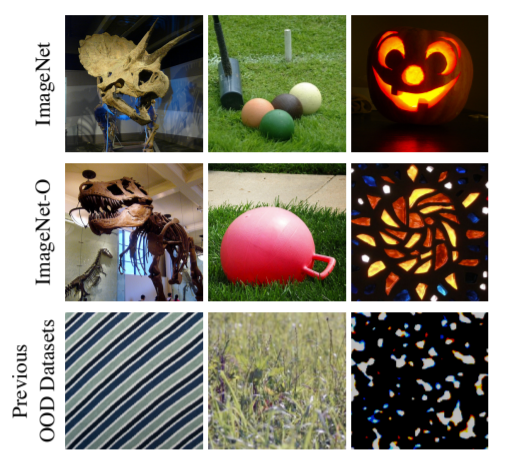}
\caption{Imagenet-O Example \cite{Hendrycks_2021_CVPR}}
\end{figure}
\subsection{Maximum Softmax Probability}
Maximum Softmax Probability (MSP) \cite{DBLP:journals/corr/HendrycksG16c} is the baseline approach OOD detection that is widely used in literature. It is a predictive score based approach where, as the name suggests, the negative of the softmax probability of the predicted class for the sample is used as the out of distribution score, i.e. the score indicating that the data sample is out of distribution (higher the score, higher the likelihood of it being from the out-distribution). It can be formulated as follows: $Score = - \max(\frac{\exp(f_i(x)}{\sum_{i}^{C}\exp(f_i(x))})$,  where $f()$ is a neural network that has been trained on the in-distribution dataset to classify C classes, and $x$ is a data sample used at inference. $f_i(x)$ denotes the neural network's predicted confidence for class $i$ where $i = 1 .. C$. We did our experiments for various choices of $f()$ as listed in \cite{Hendrycks_2021_CVPR}.
\subsection{ODIN}
ODIN \cite{liang2020enhancing} is another predictive score-based technique that uses temperature ($T > 0$) scaling of the softmax score  and addition of noise ($\epsilon > 0$) to inputs to separate the softmax distributions of in and out distribution data, and enhance the ability to use the resulting temperature scaled softmax score for more (as compared to MSP) effective OOD detection. The addition of $\epsilon > 0$ noise: $\Tilde{x} = x - \epsilon(-\nabla_x\log S_{\hat{y}}(x ; T))$, where $S_i(x ; T) = \frac{\exp(f_i(x)/T}{\sum_{i}^{C}\exp(f_i(x)/T)}$ is the temperature scaled softmax score of in-distribution class $i$ from a neural network $f()$ trained to classify $C$ classes. $\hat{y}$ denotes the class with the maximum predicted softmax score. Thus, the OOD Score computed using this technique is the maximum temperature scaled softmax score for the noise added input, i.e. $Score = -\max(S_i(\Tilde{x}; T))$ where $\Tilde{x}$ is the noise perturbed input as defined before.\\
$T$ and $\epsilon$ are hyperparemeters that need to be tuned to optimize performance for difference in-distribution and out-distribution combinations [\cite{liang2020enhancing}, \cite{tajwar2021true}]. Generally a higher value of $T$ is recommended [\cite{lee2018simple}, \cite{hsu2020generalized}], whereas performance is very sensitive to choice $\epsilon$. We also performed a hyperparameter search (within our compute limits) and have shown results for different combinations of $T$ and $\epsilon$. \textbf{Please note that this hyperparameter tuning for the best choice of $T$ and $\epsilon$ assumes prior access to OOD data, or parts of it, in order to be able search for these values that optimize OOD detection performance on the same.}
\subsection{Energy Based OOD Detection}
Energy score based OOD detection proposed in \cite{liu2021energybased} is an elevated version of the predictive score-based techniques we have seen so far. This approach can work in both settings - when we have prior access to OOD data and when we don't. In the test setting (where we don't assume prior access to OOD data) a pre-trained network can be used to get the energy score for each data sample and this energy score can be then used for OOD detection in a manner similar to the Softmax score mentioned in MSP \cite{DBLP:journals/corr/HendrycksG16c}. If prior access to OOD data is present, en energy bounded learning objective is used to fine-tune the network.\\
Energy based models return a scalar value for each input sample. This scalar value - the energy score - is lower for observed data and higher for unobserved data. With access to OOD samples in advance, one trains with  a modified objective that is in essence the standard cross entropy loss with L2 regularization defined in terms of energy. The L2 regularization term that uses energy, is present to explicitly create an energy gap between the in-distribution and out-distribution data by assigning low energy values to in-distribution data and high energy to out-distribution samples. For more details on the energy based approach, please check out \cite{liu2021energybased}.
\subsection{Normalizing Flows}
Deep generative models (DGM) seem to be a logical choice of approach for OOD detection, as they estimate distributions from the given input data and are able to generate simulations using the same. Thus, it would be reasonable to think that they would then place higher likelihoods on this input data i.e. in-distribution data, that they're estimating the density for, as compared to never seen before out-distribution data. However, as shown in \cite{zhang2021understanding} and other prior related work, DGMs assign higher likelihood to OOD data, as compared to in-distribution data. This failure of DGMs in doing OOD detection has motivated works like \cite{kirichenko2020normalizing}, \cite{maaloe2019biva} and \cite{schirrmeister2020understanding}, that have introduced modifications in DGMs for improved OOD detection. We choose to further evaluate the improvements proposed for Normalizing Flows in \cite{kirichenko2020normalizing} for our in-distribution and out-distribution setting.\\
As stated in \cite{kirichenko2020normalizing}, normalizing flows \cite{Tabak2013AFO} are a type of DGM that use invertible transformations of a base latent distribution $p_Z(z)$ in order to model a target distribution $p_X(x)$. $p_Z(z)$ is generally assumed to be a standard Gaussian. Flows are optimized for maximizing the log-likelihood of the input data by updating the parameters of the invertible transformation $f^{-1}$, which in this particular case are the coupling layers that masks some of the input and attempt to predict this masked region. The histogram of the log-likelihood placed by the flows on in-distribution data should generally be higher than the out-distribution data. A more quantitative way of evaluating OOD detection using Flows - which we will be using - is by again viewing OOD detection as a binary classification task and computing the AUROC.\\
One of the main reasons behind flows failing in OOD detection is that they learn the local image/graphical properties of the data like local pixel correlations, instead of the semantic properties of the input data \cite{kirichenko2020normalizing}. They supported this statement by showing the latent representations learnt, as well as describing how coupling layers learn from local pixel correlations and co-adapt to data from the previous coupling layers when learning to predict the masked pixels. This necessitates shifting the inductive biases of flows models to learn semantic properties of the data instead, so that likelihood is assigned based on the semantic content of the images. The improvements proposed by \cite{kirichenko2020normalizing} are summarized below - \\
\begin{enumerate}
    \item Changing objective/loss term to include a term to minimize likelihood on a specific OOD dataset, whilst maximizing likelihood on the target in-distribution dataset.
    \item Masking strategy used in the coupling layers - from checkerboard masking, to horizontal masks and cyclic masks. This is done so that masking is done in a manner that the model cannot simply use the information from the pixels around it and learn that, but instead learns to understand the image from a more global/semantic level. 
    \item Adding a bottleneck to st-networks (scale and shift transforms modelled as a network) to prevent the model from learning local pixel level relations, as the input image data has been projected into a lower dimensional space and will thus deter the model from learning local pixel level properties.
    \item Using image embeddings instead of raw image data (with pixels) to avoid learning local pixel-level properties, and forcing the network to learn the semantic properties of the in-data. 
\end{enumerate}
\textbf{The improvements to this method also require prior access to OOD data.}
\section{Experiments and Results}
Following the metrics stated in \cite{Hendrycks_2021_CVPR} and \cite{tajwar2021true}, the metrics that we will be using for evaluating the various methods are:\\
\begin{enumerate}
    \item \textbf{AUPR} - Area under the precision recall curve. Higher is better.
    \item \textbf{AUROC} - Area under the receiver operator characteristic, which for binary classification problem like - OOD detection - is a plot of the true positive rate vs the false positive rate. Higher is better.
    \item \textbf{FPR@95} - The false positive rate when the $95\%$ of true positives have been correctly classified. Lower is better.
\end{enumerate}
\textbf{The code for these experiments can be found \href{https://github.com/anugyas/OOD-on-NAE}{here.}}
\subsection{Maximum Softmax Probability}
We use pre-trained ImageNet-1K models from PyTorch \cite{NEURIPS2019_9015} and other open sourced repositories like \cite{pretrainedmodels}. Thus, the \textbf{In Distribution} is ImageNet-1K and the \textbf{Out Distribution} is ImageNet-O. We evaluated this technique for different neural network architectures. The AUROC/AUPR/FPR@95 values for OOD detection via MSP for different architectures can be seen in Figures 2,3 and 4 respectively.\\
\textbf{The results from the MSP experiments are - when using AUROC and AUPR as a metric for measuring MSP's ability to do OOD detectioon, Dual Path Network - 98 was the best performing architecture to perform OOD detection on ImageNet-O with MSP. However, when looking at the FPR@95 metric, ResNet 101 was the best performing architecture}.
\begin{figure}[t]
\includegraphics[width=8cm, height=6cm]{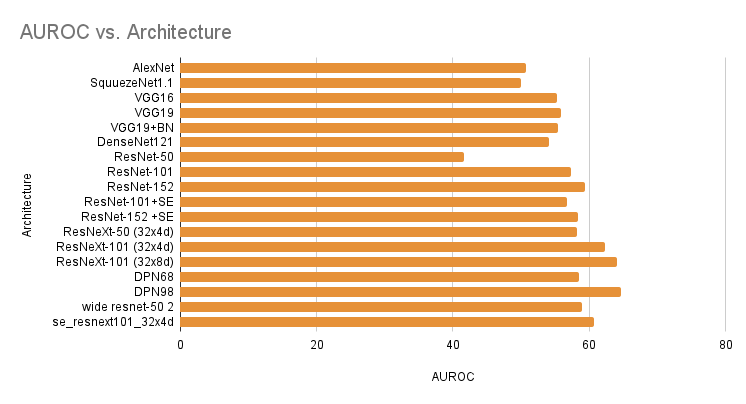}
\caption{MSP: AUROC for different model architectures}
\end{figure}
\begin{figure}[t]
\includegraphics[width=8cm, height=6cm]{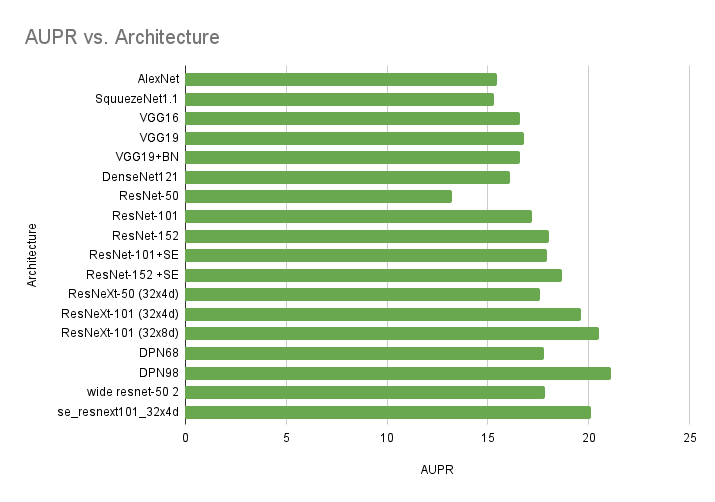}
\caption{MSP: AUPR for different model architectures}
\end{figure}
\begin{figure}[t]
\includegraphics[width=8cm, height=6cm]{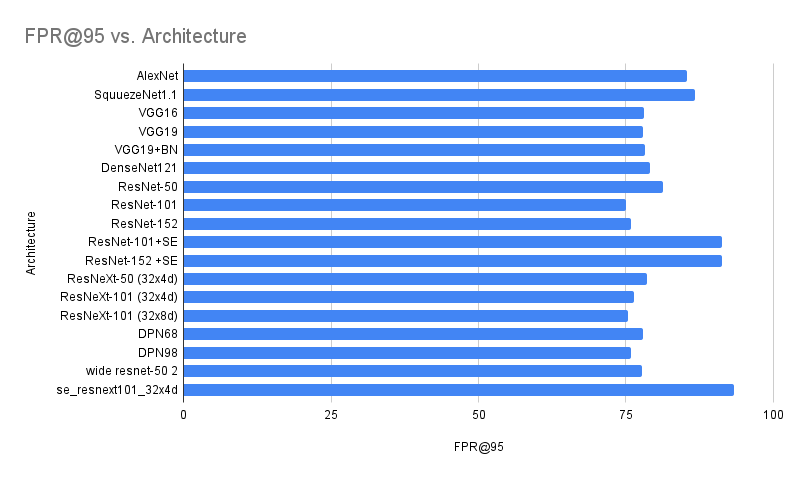}
\caption{MSP: FPR@95 for different model architectures}
\end{figure}
\subsection{ODIN}
Similar to MSP, we use pre-trained ImageNet-1K models from PyTorch \cite{NEURIPS2019_9015} and other open sourced repositories like \cite{pretrainedmodels}. Thus, the \textbf{In Distribution} is ImageNet-1K and the \textbf{Out Distribution} is ImageNet-O. We evaluated this technique for different neural network architectures. For each of the architectures, we also did a hyperparameter search for different combinations \cite{tajwar2021true} of temperature $T = [100, 1000]$ and noise $\epsilon = [0.001, 0.0014, 0.002, 0.0024, 0.005]$. The baseline value of $T=1000$ and $\epsilon=0$, in accordance with \cite{tajwar2021true}.The AUROC, AUPR and FPR@95 plots for ODIN can be seen in Figures 5, 6 and 7 respectively.\\
\textbf{The results of the ODIN experiments are - the highest AUROC - 82.18\% was with ResNet-152 with $T=100$ and $\epsilon=0.0024$ ; the highest AUPR - 60.25\% was with VGG-16 with $T=1000$ and $\epsilon=0.0024$ ; the lowest FPR - 76.93\% was with ResNext 101 32x8d with $T=1000$ and $\epsilon=0$.}\\
\textbf{\textit{On comparing with MSP's best AUROC and AUPR, ODIN has certainly outperformed MSP on ImageNet-O in our experiments.}}
\begin{figure}[t]
\includegraphics[width=9cm, height=6cm]{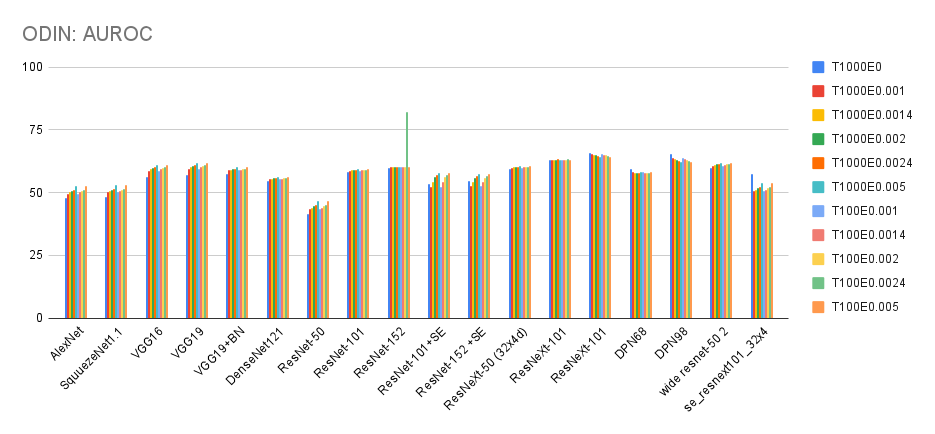}
\caption{ODIN: AUROC for different model architectures and different values of $T$ and $\epsilon$ - denoted as E in the plot}
\end{figure}
\begin{figure}[t]
\includegraphics[width=9cm, height=6cm]{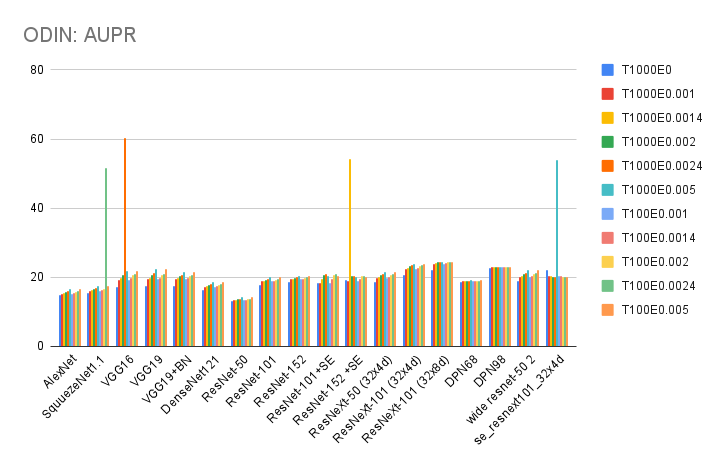}
\caption{ODIN: AUPR for different model architectures and different values of $T$ and $\epsilon$ - denoted as E in the plot}
\end{figure}
\begin{figure}[t]
\includegraphics[width=8cm, height=6cm]{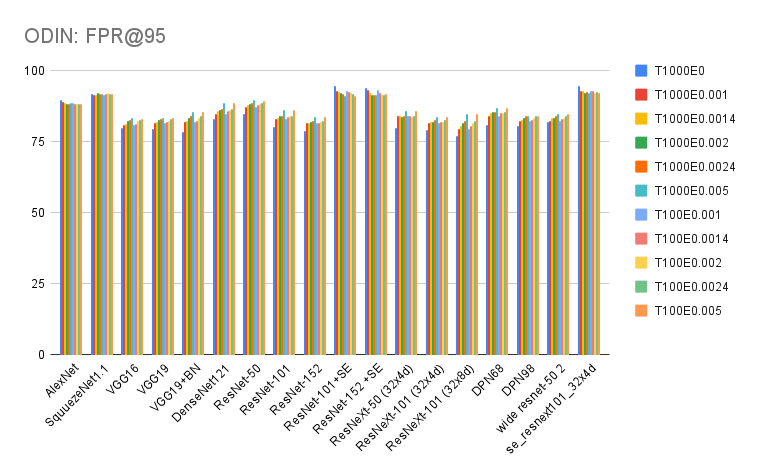}
\caption{ODIN: FPR@95 for different model architectures and different values of $T$ and $\epsilon$ - denoted as E in the plot}
\end{figure}
\subsection{Energy Based OOD Detection}
For the energy based experiments -  we fine tuned a CIFAR-100 pre-trained WideResNet on TinyImagNet, which thus makes up the in-distribution and also used some OOD samples from ImageNet-O to help increase the energy gap between in-distribution and out-distribution points. The results can be seen in Table 1.
\begin{table}[!htbp] \centering 
\caption{OOD Detection Performance on ImageNet-O}{
\begin{tabular}{|p{2cm}|p{1cm}|p{1cm}|p{0.7cm}|}
\hline
\hline
AUROC & AUPR & FPR@95\\
\hline
70.35\% & 13.06\% & 87.97\% \\
\hline
\hline
\end{tabular}
}
\end{table}
\subsection{Normalizing Flows}
The experiments for Normalizing Flows were done with resized 32 x 32 images. Tiny Image Net \cite{TinyImageNet} was used for all training purposes and thus forms the in-distribution. The out-distribution is the ImageNet-O dataset. As discussed in the methods, there were 4 proposed improvements to Normalizing flows in \cite{kirichenko2020normalizing} for improving OOD detection performance. The result of the baseline RealNVP with a ResNet based st-network can be seen in Figure 8. The AUROC for the baseline is around 78\%.  We tried the following experiments in accordance with the proposed improvements.
\begin{enumerate}
    \item Using the updated objective in \cite{kirichenko2020normalizing} that uses some OOD samples while training - Equation 7 in \cite{kirichenko2020normalizing}. The results can be seen in Figure 9. The AUROC is 74\%. 
    \item Using the updated objective mentioned above along with Cycle-masks. The result can be seen in Figure 10. The AUROC is 70\%. 
    \item Using image embeddings instead of raw image data. The results can be seen in Figure 11. The training would collapse after a approximately 50 epochs. We tried many learning rates as well as other hyperparameter choices and only reducing the learning rate stabalized it for longer but it would then eventually collapse and result in NaNs. The results after approximately 50 epochs can be seen in Figure 10.
\end{enumerate}
\begin{figure}[t]
\includegraphics[width=9cm, height=6cm]{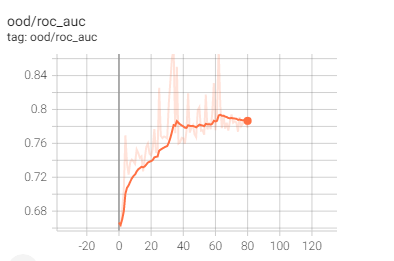}
\caption{Flows: AUROC for RealNVP baseline}
\end{figure}
\begin{figure}[t]
\includegraphics[width=9cm, height=6cm]{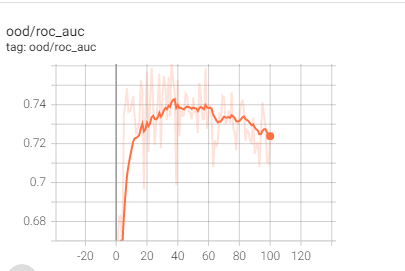}
\caption{Flows: AUROC for RealNVP trained with a subest of OOD samples}
\end{figure}
\begin{figure}[t]
\includegraphics[width=9cm, height=6cm]{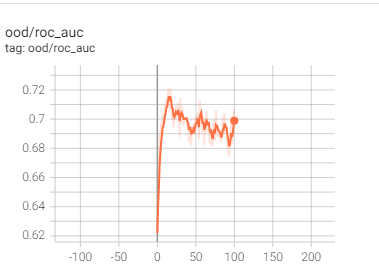}
\caption{Flows: AUROC for RealNVP trained with a subset of OOD samples and Cycle Mask}
\end{figure}
\begin{figure}[t]
\includegraphics[width=9cm, height=6cm]{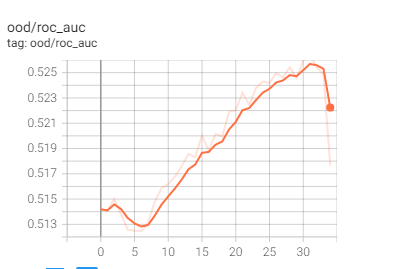}
\caption{Flows: AUROC for RealNVP trained on image embedding}
\end{figure}
\section{Conclusion}
\textbf{From all the experiments that we did, the results from ODIN were the best.}
As shown in \cite{tajwar2021true} and \cite{zhang2021understanding}, what defines the in-distribution and out-distribution is crucial in the choice of approach and the resulting performance. Our results are also consistent with that view-point where changing the in-distribution from ImageNet-1K to TinyImageNet severely impacted the performance, and also revealed some inconsistencies in the improvements suggested in approaches like \cite{kirichenko2020normalizing}.\\
The results above also make it evident that there is a lot of variability in the performance of each of these techniques, from choice of architecture - as shown in MSP and ODIN, to choice of hyperparameters - like temperature $T$ and $\epsilon$ for ODIN, to choice of including OOD samples during training or not and more.\\
Given our experiments, training on the complete ImageNet 1-K dataset is the most important factor in the success of any of these OOD detection techniques for ImageNet-O. Even when evaluating approaches that have been trained/fine-tuned only on the TinyImageNet dataset, the steps for improvement that worked for smaller in-distribution, out-distribution combinations, seems to be not be as effective for the novel ImageNet-O.
\section{Future Scope}
As our work is one of the first of its kind - benchmarking OOD detection for ImageNet models on the novel ImageNet-O dataset, and also the general lack of work on OOD detection with in-distribution as ImageNet - most focus on CIFAR-10 and CIFAR-100,  some interesting future directions that we think this work can take are:
\begin{enumerate}
    \item Using the complete ImageNet-1K dataset for Methods 3 and 4, with enough compute and time on our side
    \item More comprehensive hyperparameter tuning for $T$ and $\epsilon$ for ODIN, as they seem to play a critical role in ODIN's performance, regardless of depth - VGG16 was one of the best performing ones!
    \item Varying the percentage of OOD samples that are used when they are used in training for the energy based and normalizing flows approaches. Additionally, the same can be done when using OOD samples for hyperparameter tuning $T$ and $\epsilon$ for ODIN.
    \item Applying and evaluating combinations of improvements proposed in \cite{kirichenko2020normalizing}
    \item Trying more architectures with the energy based technique
    \item Trying more variety of st-networks in the normalizing flows approach
    \item Exploring more masking techniques for normalizing flows
    \item Trying different architectures for getting image embeddings to train normalizing flows on
\end{enumerate}
{\small
\bibliographystyle{ieee}
\bibliography{egbib}
}
\end{document}